\pdfoutput=1

\documentclass[11pt]{article}

\usepackage{ACL2023}

\usepackage{times}
\usepackage{latexsym}

\usepackage[T1]{fontenc}

\usepackage[utf8]{inputenc}

\usepackage{CJKutf8}

\usepackage{microtype}

\usepackage{inconsolata}

\usepackage{booktabs}
\usepackage{latexsym}
\usepackage{listings}
\usepackage{enumitem}
\usepackage{graphicx}

\usepackage{CJKutf8}

\usepackage{array}
\usepackage{float}
\usepackage{multirow}

\usepackage{xstring}

\newcommand{\dividepage}{%
  \par\noindent
  \rule{\linewidth}{.4pt}{\dotfill}%
  \par
}

\usepackage{color,colortbl}
\definecolor{gray}{rgb}{0.75, 0.75, 0.75}

\definecolor{princetonorange}{rgb}{1.0, 0.56, 0.0}
\usepackage{xcolor}
\usepackage{hyperref}
 \definecolor{darkblue}{rgb}{0, 0, 0.5}
  \hypersetup{colorlinks=true, citecolor=darkblue, linkcolor=darkblue, urlcolor=darkblue}

\usepackage{fdsymbol}

\usepackage{cleveref}

\usepackage{pifont}

\usepackage{arabtex}
\usepackage{utf8}

\setcode{utf8}

\usepackage{utfsym}
\newcommand{\newcrossmark}{\scalebox{0.75}{\usym{2613}}}

\def\Plus{\texttt{+}}

\font\tenipa=tipa10
\def\schwa{{\tenipa\char64}}

\title{MASSIVE Multilingual Abstract Meaning Representation:\\ A Dataset and Baselines for Hallucination Detection}

\author{
    Michael Regan\textsuperscript{\dag},
    Shira Wein\textsuperscript{\ddag},
    George Baker\textsuperscript{+},
    Emilio Monti\textsuperscript{*} \\
    \textsuperscript{\dag}Paul G. Allen School of Computer Science \& Engineering \\
    \textsuperscript{\ddag}Georgetown University \\
    \textsuperscript{+}University of Colorado Boulder \\
    \textsuperscript{*}Amazon, Cambridge, UK \\
    \texttt{mregan@cs.washington.edu,monti@amazon.co.uk}
}

\begin{document}
\maketitle

\begin{abstract}
Abstract Meaning Representation (AMR) is a semantic formalism that captures the core meaning of an utterance. There has been substantial work developing AMR corpora in English and more recently across languages, though the limited size of existing datasets and the cost of collecting more annotations are prohibitive. With both engineering and scientific questions in mind, we introduce MASSIVE-AMR, a dataset with more than 84,000 \emph{text-to-graph annotations}, currently the largest and most diverse of its kind: AMR graphs for 1,685 information-seeking utterances mapped to 50\Plus{ }typologically diverse languages. We describe how we built our resource and its unique features before reporting on experiments using large language models for multilingual AMR and SPARQL parsing as well as applying AMRs for hallucination detection in the context of knowledge base question answering, with results shedding light on persistent issues using LLMs for structured parsing.
\end{abstract}

\section{Introduction}
\noindent
Knowledge base question answering (KBQA) has a long history in natural language processing, with the task of retrieving an answer from a knowledge base such as Wikidata or DBPedia \citep{Lehmann2015} integral to many large-scale question answering systems \citep{Kapanipathi2021}. In KBQA, a question is converted into a structured query language such as SPARQL, an executable semantic parse. However, data to train models is expensive, few multilingual resources are available, and performance is limited for long-tail queries, a problem compounded by arbitrary variability in form-meaning mappings across languages \citep{Croft2002}.

Most notably, research in multilingual KBQA is hindered by lack of data \citep{Usbeck2018,Cui2022,Perevalov2022}. Following work using meaning representations for this problem, we create a dataset 20 times larger and with 5-6 times more languages than existing resources \citep{Lee2022} (\Cref{tab:statistics}). For MASSIVE-AMR, we select 1685 QA utterances with manual translations from MASSIVE
\citep{Fitzgerald2023} and manually compose Abstract Meaning Representation (AMR) graphs \citep{Banarescu2013}, amounting to 84,000 text-to-graph annotations, a significant boon to AMR and KBQA research.

\begin{table}[tb]
    \centering
    \small
    \begin{tabular}{l|l|l|l}
    \toprule
    {} & \scriptsize\textbf{AMR3.0} & \scriptsize\textbf{QALD9-AMR} & \scriptsize\textbf{OURS} \\
    \midrule
    \# of languages & 1 & 9+ & 52 \\
    domain & various & QA & QA \\
    \# utterances & 59K & 508 & 1685 \\
    \# utts-to-graphs & 59K & {5}K & {84}K \\
    mean tokens/utt  & 15.9 & EN: 7.5 & EN: 8.2\\
    entities & - & not local  & local  \\
    gold SPARQL & No & Yes & No \\
    \bottomrule
    \end{tabular}
    \caption{Other AMR treebanks and ours, MASSIVE-AMR. Compared with QALD9-AMR \citep{Lee2022}, MASSIVE-AMR covers more languages, has more utterances, and has localized or translated entities for each language (see exs. \Cref{tab:QALD9}).}
    \label{tab:statistics}
\end{table}

Graphs with localized entities (\Cref{tab:QALD9}) and the long-tail utterances in MASSIVE-AMR (Appendix \ref{appendix:long-tail}) increase the challenge of our dataset (\S\ref{sec:corpus_creation}).
To explore the resource's utility, we design and carry out experimentation leveraging AMRs to gauge model confidence in SPARQL query production (\S\ref{sec:task}), reporting on multilingual structured parsing and SPARQL relation hallucination detection using large language models (LLMs) (\S\ref{sec:experiments}).

Our research contributions thus include: (1)~creation of the largest-scale multilingual AMR question corpus to date; (2)~evaluation of LLMs on parsing of SPARQL and AMRs structures across languages; and (3)~design, development, and evaluation of generative models leveraging AMRs for SPARQL relation hallucination detection.\footnote{We release the MASSIVE-AMR training and validation data at \url{https://github.com/amazon-science/MASSIVE-AMR}.}

\begin{table}
    \centering
    \footnotesize
    \begin{tabular}{c l|l}
    \toprule
    {} & \textbf{Utterance} & \textbf{AMR} \\
    \midrule
     \parbox[t]{2mm}{\multirow{9}{*}{\rotatebox[origin=c]{90}{\textbf{MASSIVE-AMR}}}}  & when was \underline{obama} born & {(b / bear-02} \\
     {} & {} & \-\ \-\ :ARG1 (o / ``obama") \\
     {} & {} &  \-\ \-\ :time (u / unknown)) \\
     {} & {} & {} \\
      {}  & quand est né \underline{sarkozy} & (b / bear-02 \\
      {} & {} & \-\ \-\ :ARG1 (s / ``sarkozy") \\
     {} & {} &  \-\ \-\ :time (u / unknown)) \\    
      {} & {} & {} \\
      {} & +50 langs. & +50 AMRs, local entities \\
      \midrule
      \parbox[t]{2mm}{\multirow{5}{*}{\rotatebox[origin=c]{90}{\textbf{QALD9-AMR}}}}    & Who developed \underline{Skype}? & {(d / develop-02}\\
       {} & Qui a développé \underline{Skype}? &  \-\ \-\  {:ARG0 (u / unknown)} \\
       {} & {} &  \-\ \-\  :ARG1 (s / ``Skype")) \\
       {} & {} & {} \\
       {} & 9+ langs. & {Same AMR, all langs.} \\
    \bottomrule
    \end{tabular}
    \caption{MASSIVE-AMR (top) has localized entities (English-US `obama', French-FR `sarkozy') and covers >5x more languages compared to QALD9-AMR (bottom). AMRs simplified to fit table.}
    \label{tab:QALD9}
\end{table}

\section{Related Work}
\label{sec:related_work}
\noindent
We present related work in QA, Knowledge base question answering (KBQA), the AMR formalism, AMRs for KBQA, and hallucination detection.

\subsection{Question Answering}
\noindent
Question answering (QA) is the task of retrieving or predicting an answer to a natural language query given document(s), a list of answers, knowledge triples, or with a generative model. QA encompasses research in Information Retrieval \citep{Lewis2020}, Machine Reading Comprehension (MRC) \citep{Das2018}, and Open-Domain Question Answering \citep{Lewis2021,Zhang2023}. Research  targeting model confidence for calibration of QA systems \citep{Jiang2021,Kadavath2022} has aims similar to our own.

For research in multilingual dialogue systems, MASSIVE~\citep{Fitzgerald2023} is a collection of 20K utterances with manual translations into 50+ typologically diverse languages (with 52 languages in v1.1). For our dataset, we select all QA utterances from  MASSIVE and add AMR annotations (see \Cref{sec:corpus_creation}).

\subsection{Knowledge Base Question Answering}

\noindent
Knowledge base question answering (KBQA) is the task of retrieving answers from a knowledge base given a  question. The challenges in retrieving textual information are fundamentally different from the primary challenge of KBQA: producing semantically accurate knowledge base queries. 

Various approaches to KBQA have been proposed over the decades, including converting queries to logical forms, semantic parses, and decomposing complex questions \citep{Zelle1996,Zettlemoyer2005,Talmor2018}. Scalable KBQA systems utilize structured representations (SPARQL) to query a knowledge base (e.g., DBPedia\footnote{\url{https://www.dbpedia.org/}}), a collection of triples of form <subject, rel$_j$, object> with rel$_j$ a semantic relation from ontology $\mathcal{R}$ (of various sizes, e.g., $|\mathcal{R}_{\mathrm{DBPedia}}|>2500$). Baselines for SPARQL parsing are available \citep{Banerjee2022}, with a central challenge being how to identify parsed queries not covered by a given $\mathcal{R}$, cases where models tend to hallucinate relations.

In the age of large language models, querying manually-curated knowledge bases provides numerous advantages such as: (1)~factuality guarantees, (2)~the ability to update information in real time, and (3)~risk mitigation for users, reducing exposure to sensitive or toxic content. With these motivations in mind, we turn our attention to AMRs.

\subsection{Abstract Meaning Representation}

\noindent
Abstract Meaning Representation (AMR) \citep{Banarescu2013} is a linguistic formalism that represents utterance meaning as directed, mostly acyclic graphs. Graph nodes denote key concepts associated with the meaning of the utterance, targeting events and event participants. Nodes in turn are connected by labeled edges for event-event, event-entity, entity-entity, and other relations.

Early AMR research focused on text-to-AMR parsing, with the JAMR parser \citep{Flanigan2014} paving the way for state-of-the-art models based on transitions \citep{Drozdov2022}, seq2seq approaches \citep{Bevilacqua2021}, and ensemble distillation \citep{Lee2022}. In lieu of such heavily engineered approaches, we target generative models with in-context learning and fine-tuning following recent work \citep{Ettinger2023}.

The original AMR reference-based metric is Smatch \citep{Cai2013}, a measure of overlapping triples, which has led to the newly optimized Smatch++ \citep{opitz-2023-smatch} and S2match \citep{s2match/tacl_a_00329} which uses embeddings to match concepts within triples. \Citet{Wein2022} released multilingual AMR metrics such as XS2match using LaBSE embeddings \citep{feng2020language} for cross-lingual AMR evaluation.

AMRs were not designed to function across languages \citep{Banarescu2013}, and while language has a measurable effect on AMR structure \citep{wein-etal-2022-effect}, efforts have been made to effectively represent the meaning of non-English sentences in AMRs \citep{Xue2014,hajic-etal-2014-comparing,10.1162/coli_a_00503}. In  typology, a Uniform Meaning Representation \citep{Gysel2021} helps account for formal and semantic differences across languages more consistently than AMR, and work tying multilingual resources to a common formalism is ongoing \citep{Navigli2022}.

\subsection{AMR for KBQA}
\noindent
Using symbolic representations for QA is well studied in NLP~\citep{niu2023bridging,wang2023exploiting}. A mapping of AMR nodes to SPARQL concepts and variables is shown to improve KBQA systems~\citep{Kapanipathi2021}, and sequence-to-sequence models learn to apply these rules selectively for improved generalization \citep{Bornea2022}. 

The multilingual QA resource most similar to ours is QALD9-AMR \citep{Lee2022}, which maps utterances from 9+ languages to the same English-only AMR and gold SPARQL queries~\citep{Usbeck2018}. In comparison, graphs in MASSIVE-AMR consist of multilingual entities (\Cref{tab:QALD9}) either translated or localized (e.g., a regional entity for where the language is spoken) for each of 50+ languages (\Cref{tab:QALD9,tab:examples}).

\subsection{Hallucination detection}
\noindent
\textit{Hallucinations}, the inclusion of flawed or incongruous assertions in synthetic text, represent a persistent problem with LLMs \citep{Ji2023}. Much research in hallucination detection targets the \textit{text-to-text} paradigm, for example checking factuality or faithfulness of summarized texts \citep{Gabriel2021,Qiu2023} or proposing mitigation strategies to make synthetic text attributable \citep{Aksitov2023,Rashkin2023}. In contrast, we examine \textit{text-to-graph} systems that produce executable semantic parses, experimenting with AMRs to detect \textit{easy} and \textit{hard} cases of \textit{semantic relation hallucination}, ranking parses of dual representation types in a joint space, as we will detail in \Cref{sec:task}.

\section{Data: Corpus Creation}
\label{sec:corpus_creation}
\noindent
To create a corpus of multilingual AMR graphs, we started with an existing dataset of QA utterances, tailored AMR 3.0 guidelines to our use case, trained a team of professional annotators to create AMRs for English utterances, and then made automatic mappings to multilingual utterances using existing entity mention spans, a process which from start to finish took three months. In this section, we report details about the data we started with, guidelines, and annotation agreement scores.

 \textbf{Acquiring scaleable multilingual data}. We wanted a resource targeting a wide distribution of QA utterances and thus selected 1685 English examples from MASSIVE~\citep{Fitzgerald2023} including entity annotations like in the multilingual examples in \Cref{tab:examples}.

\begin{table}[h]
    \small
    \centering
    \begin{tabular}{r|l}
    \toprule
    \textbf{Lang.} & \textbf{Example utterance} \\
    \midrule
     en-US & what is the population of {[}place: new york{]} \\
 sl-SL & koliko prebivalcev ima {[}place: ljubljana{]} \\
 it-IT & qual è la popolazione di {[}place: roma{]} \\
 sq-AL & cila është popullësia e {[}place: tiranës{]} \\
 cy-GB & beth yw poblogaeth {[}place: efrog newydd{]} \\
 af-ZA & wat is die bevolking van {[}place: kaapstad{]} \\
 is-IS & hver er íbúafjöldi {[}place: reykjavíkur{]} \\
 az-AZ & {[}place: sumqayıtın{]} \schwa halisi n\schwa q\schwa d\schwa rdir \\
\bottomrule
    \end{tabular}
    \caption{Example multilingual questions from MASSIVE \citep{Fitzgerald2023} about the populations of  regional cities, with annotations for entity spans and types given.}
    \label{tab:examples}
\end{table}

\textbf{Long-tail QA}. Many utterances in MASSIVE are described as long-tail, that is, associated with low user feedback in interactions with a digital assistant. In some cases, it is clear what increases friction (an incomplete utterance, or a speech-to-text error). Examining translations of English utterances provides insight (\Cref{appendix:long-tail}).

\textbf{Localized entities}. In comparable datasets \citep{Cui2022,Perevalov2022}, entities are shared across languages (e.g., English \emph{Where did Abraham Lincoln die?} corresponds to German \emph{Wo starb Abraham Lincoln?}). To address challenges of large-scale QA, MASSIVE entities are mostly language-specific, e.g. German questions target German entities (\emph{wo starb otto von bismarck}\footnote{MASSIVE utterances are uncased with no punctuation.}).

\textbf{AMR datasets differ in composition}: AMR 3.0~\citep{Banarescu2013} is based on news and other written discourse and consists of relatively few factoid or information-seeking questions (less than 10\%). In contrast, MASSIVE-AMR includes requests about currency conversions, quantities, comparative and superlatives, and simple arithmetic. For more details about how the corpora compare, see \Cref{appendix:massive} and \Cref{tab:top-nodes}.

\textbf{Annotation principles: Canonical forms}. In keeping with original AMR guidelines, an AMR captures meaning, not form \citep{AMRGuidelines}. We hence prefer canonical forms for utterances like currency conversion and arithmetic: e.g., `how much is the euro versus the dollar' and `what is the euro worth compared to the dollar' map to similar graphs. Likewise, arithmetic questions are associated with top node `equal-01' even without token `equal' present (`how much is two plus two' and `sum of two and two' treated like `what does two and two equal'). 

\textbf{Question-imperative continuum}. It proved difficult to reach agreement for annotations of question versus imperative forms. In English, `could you tell me the price of google', `what is the price of google', and `tell me the price of google' share the same meaning. However, treating the imperative (e.g., an embedded question `tell me what the price is') as a question is out-of-line with AMR 3.0. The guideline we adopt is to preserve imperative form and treat polite questions (e.g., English `could you tell me the price') the same as  base question forms (e.g., `what is the price').

\textbf{Annotation agreement scores}. 4-5 trained annotators created AMRs for 1685 utterances, examining differences in batches of 200 weekly, with inter-annotator agreement ranging from 78-82\% Smatch, comparable to reported agreement for AMR experts \citep{Banarescu2013}. We note that MASSIVE-AMR consists of many similar questions and simple utterances, with on average 50\% fewer tokens compared to AMR 3.0 (\Cref{tab:statistics}). We select the single best AMR in candidate sets and manually retrofit to increase consistency.

\textbf{For non-English entities}, we replace AMR node labels using MASSIVE annotations. We note that not all utterances have annotations, and that a lack of entity alignments adds noise since often word order matters (e.g., currency conversion). To improve data quality, we manually curate validation and test sets (25\% of total).

\section{SPARQL Hallucination Detection}
\label{sec:task}
\noindent
Our original motivation for creating a multilingual AMR dataset (\S\ref{sec:corpus_creation}) was to help improve large-scale QA systems. Scaleable QA systems often utilize structured representations (e.g., SPARQL) for knowledge base retrieval, pairing a natural language utterance with an executable semantic query. The SPARQL in the Wikidata or DBPedia case is straightforward: we get a question in, the system produces an answer out. However, in practice we simply need a system capable of judging if a given answer is correct, which using generative methods we study as \textit{hallucination detection}.

\begin{figure*}[t]
    \centering
    \includegraphics[scale=0.58]{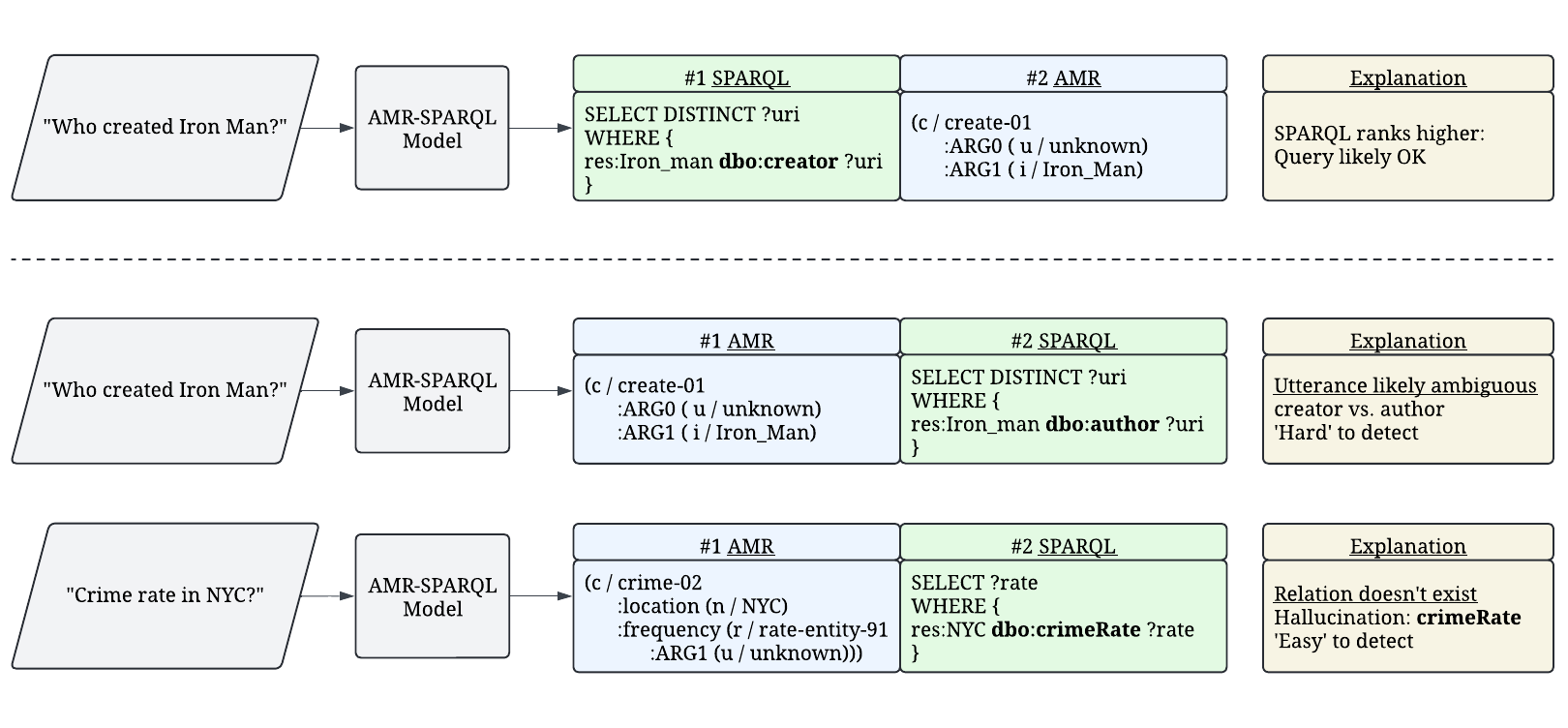}
    \caption{As a proxy for QA correctness, we test a joint AMR-SPARQL model, controlling for semantic relations (in \textbf{bold}). Given an {utterance} like \textit{Who created Iron Man?}, a model outputs a \texttt{N}-best list of candidates of mixed representation types. When the relation \textbf{creator} is allowed (top), we expect the model to rank SPARQL higher than AMR. If the we change the ontology, the AMR may rank higher (middle), suggesting an ambiguity exists (\textbf{creator} $\approx$ \textbf{author}). Models also produce non-existent relations (bottom), detected via ranking or a look-up operation. 
}
    \label{tab:joint_model}
\end{figure*}

\textbf{Hallucinations}. A problem in open-domain question-answering regards \textit{hallucinations}, cases when effectively the target Ontology (in our case, DBPedia) does not have valid symbols for a given input question (see \Cref{tab:joint_model}). For example, if the relation `crimeRate' does not exist, a SPARQL generation model may stumble on a question like `What is the crime rate in LA?' by parsing a query with a non-existing relation, which we can verify with a set membership check. A harder case to detect is when the model predicts a relation for an utterance that is ambiguous, e.g., `Who created Iron Man' may refer to its fictional (Tony Stark) or non-fictional (Stan Lee) creator. We would like to design and test methods for the detection of such cases using LLMs.

An advantage of AMR is that its ontology is open: i.e. if a given concept is missing, we can practically lemmatize the English. Or more often, AMR tends to be more granular, and more complex meanings (that in an Ontology might be collapsed into a single symbol) are split into several constituents (i.e. `crimeRate' might be a single symbol in an Ontology, but it is instead split into constituents by AMR). Hence, hallucinations are much less of a problem in AMR.

We hypothesize that if we train a single semantic parser to parse both SPARQL and AMRs, simply mixing the training data (i.e. for multi-task learning), and produce multiple parse candidates in a target \texttt{N}-best, the inclusion of AMRs will allow us to detect SPARQL hallucinations. That is to say, a high confidence AMR and lower confidence SPARQL serve as a signal that a given utterance is not covered by an ontology or is in some way ambiguous, as in the examples in \Cref{tab:joint_model}.

We examine dual subtasks of SPARQL hallucination detection: (1) How accurate are models at the \textbf{easy} task of checking \textit{set membership}, in our case, verifying produced relations are in a given relation set:
$$r_{pred} \stackrel{?}\in \mathcal{R}_{given}$$ 
\noindent
and, (2) How good are models at flagging ambiguous queries (e.g., `Who created Iron Man?'), the task of \textbf{hard} hallucination detection, detailed more in the next section.

\section{Experiments}
\label{sec:experiments}
\noindent
To gain insight into our hypothesis that AMRs can help detect SPARQL relation hallucinations (\S\ref{sec:task}), we first report on experiments in semantic representation parsing, a first-of-its-kind in a diverse multilingual setting. Next, we experimentally confirm models do indeed hallucinate relations, before moving on to our target task of hallucination detection.  We compare in-context learning and fine-tuned LLMs, training and evaluating on an existing corpus of questions with gold AMRs and SPARQL and sampled MASSIVE-AMR. We are guided by the following \textbf{research questions}:
\begin{enumerate}[itemsep=0pt]\setlength{\itemindent}{0.5em}
    \item\label{rq1} How effective are LLMs at parsing AMRs and SPARQL queries across languages?
    \item\label{rq2} How prevalent are SPARQL relation hallucinations with generative models?
    \item\label{rq3} How accurate are models at detecting hallucinated SPARQL relations?
    \item\label{rq4} Can we use a joint AMR-SPARQL model to do better relation hallucination detection?
\end{enumerate}

The standard approach to study the coverage of a set of relations is use all the data associated with a relation set $\mathcal{R}$ to train semantic parser $SP_{\mathcal{R}}$; we then remove all examples that contain relation $r_j$ and train $SP_{\{\mathcal{R} - {r_j}\}}$, measuring how well the model does for queries likely to require $r_j$.

An advantage of training a joint AMR-SPARQL model from scratch is having complete control over the input relations; a disadvantage is that, in the case we use a LLM, we have no knowledge about what relations the model may have seen in pre-training. For our early experiments, we use LLMs trained on 1000s of examples without hard constraints on allowed relations\footnote{Ideally, this could be done at decoding time, setting logits of all non-relation tokens to \texttt{-inf} after a colon, an unambiguous signal of a SPARQL relation.}.

We define \textit{hallucination detection} as the ability of an LLM to verify produced relations are members of a predefined set. We consider cases of \textit{hard hallucination detection}, when a model produces a relation that may be imprecise, a case which occurs when the needed relation for a query is not covered by a given $\mathcal{R}$. For experiments, we compare in-context learning with fine-tuned LLMs.

\subsection{In-context Learning}
\noindent
For in-context learning, we use GPT models \citep{openai2023GPT4} (GPT-3.5/GPT-4-0613) with prompts of length <2400 tokens (see \Cref{appendix:prompt}) composed employing strategies we describe in this section.

\textbf{Strategy \#1: Constrain and verify relations}. Prompts include a list of allowed SPARQL relations with which we instruct the model to verify predicted relations. For in-context learning, we include eight examples of joint AMR-SPARQL predictions, with example hallucinations.

\textbf{Strategy \#2: Simulate missing relations}. To control for relations (\Cref{tab:relations}), we count DBPedia relations in QALD9-AMR training data, select the 140 more frequent relations, and set aside 1+ relations for utterances in prompt where the model should prefer AMR over SPARQL, ensuring examples abide by constraints. We define the more frequent relations as being observed >1 times, which is the case for about 50\% of the data.

\begin{table}[t]
    \centering
    \footnotesize
    \begin{tabular}{l|l}
    \toprule
    \textbf{Relations} & \textbf{Subset descriptions} \\
    \midrule
All observed         & $\mathcal{R_{\mathrm{obs}}}$ \\
In-context         & $\mathcal{R}_{\mathrm{context}} \subset \mathcal{R}_{\mathrm{obs}}$ \\
Subsets similar    & $\{\mathcal{R}^{\mathrm{sim}}_1,\dots,\mathcal{R}^{\mathrm{sim}}_j\}$,  $\mathcal{R}^{\mathrm{sim}}_i \subset \mathcal{R_{\mathrm{obs}}}$ \\
Controlled          & $r_{\mathrm{cntl}} \in \mathcal{R}^{\mathrm{sim}}_i, \notin \mathcal{R}_{\mathrm{context}}$ \\
Ground truth & $\{r_{m},..,r_{\mathrm{cntl}},..,r_{n}\} \subset \mathcal{R_{\mathrm{obs}}}$ \\
    \bottomrule
    \end{tabular}
    \caption{Different subsets of relations, $\mathcal{R}$, for experimentation. To test if a generation model adheres to instructions for allowed relations, we disallow one relation from a subset of similar relations as a control (4th row). We observe model performance for questions with ground truth relations we control (last row).}
    \label{tab:relations}
\end{table}

To test our \textit{hard hallucination detection} hypothesis, we determine DBPedia relations to control for by manually grouping similar relations (e.g., `creator,' `writer,' and `developer' are similar; \Cref{tab:relations}, row 3) and select questions associated with any of these relations. We compare predictions allowing all relations versus the allowed list less the controlled relation (\Cref{tab:relations}, row 4).

\textbf{Strategy \#3: Simulate ranking}. We would like the model to rank without access to ground truth confidence scores, so we assign random confidence scores to in-context examples using a Dirichlet distribution (K=3), dropping the minimum value.\footnote{The minimum value represents the probability density of bottom predictions in latent \texttt{N}-best ranking.} 
However, at decoding we consider only relative ranking, leaving a  rigorous examination of confidence scores for future work.

\textbf{Strategy \#4: In-context examples of hallucination detection}. Prompts (\Cref{appendix:prompt}) include cases of easy and hard hallucination detection, and we direct the model to specific cases where AMRs should rank higher.\footnote{The prompt reads: ``Rank AMRs higher when predicted SPARQL is likely wrong, like in examples 5 and 8.''} 

\subsection{Additional Controls}
\noindent
We include results with an oracle, in which we direct the model's attention to the disallowed relation, providing an upper bound on achievable performance and giving insight into analysis. For consistency across datasets, we normalize all utterances (lower case, no punctuation).

\subsection{Data: Language Subsets for Parsing}
\label{ssec:languages}

For experiments in AMR and SPARQL parsing, we identify  a subset of languages: for comparison with QALD9, we select Indo-European languages from MASSIVE-AMR, the subset we refer to as \textbf{MASSIVE-}, and a more diverse sample with different scripts and less representation in Wikipedia, referred to as \textbf{MASSIVE+} (\Cref{tab:experiment-languages}).

For structured parsing experiments using in-context learning, we sample about 100 utterances each from QALD9, MASSIVE-, and MASSIVE+ (e.g., the same 16 questions in 6 different languages), reporting average results across languages in each subset.

\subsection{Fine-tuning}
\noindent
We fine-tune joint AMR-SPARQL models using publicly available LLMs: GPT-2-XL$_{\textsc{distill}}$, a 1.5B parameter variant distilled on graph-structured knowledge \citep{West2022} and LLaMA-13B \citep{Touvron2023llama}; for model fine-tuning details, consult \Cref{appendix:models}. For a challenging test set, we select same-sized samples from QALD9 and MASSIVE-AMR (900 each) of the same Indo-European languages (namely: English, Spanish, German, French, and Russian).

\begin{table}[t]
    \small
    \centering
    \begin{tabular}{c|l|r|r|r}
    \toprule
    {} & \textbf{Language} & \textbf{\# speakers} & \textbf{\# Wiki pgs} \\
    \midrule

\parbox[t]{1mm}{\multirow{7}{*}{\rotatebox[origin=l]{90}{\scriptsize{QALD9}/\textbf{\small{MASSIVE-}}}}} & English & 1.5b & 58.7m \\
{} & French  & 320m & 12.6m \\
{} & Russian  & 258m & 7.7m \\
{} & German  & 76.5m & 7.8m \\
{} & Italian & 66m & 7.7m \\
{} & Lithuanian & 2.8m & 0.5m \\
{} & {} & {} & {} \\
    \midrule
\parbox[t]{2mm}{\multirow{8}{*}{\rotatebox[origin=l]{90}{\textbf{MASSIVE+}}}}
& Vietnamese  & 85.2m & 19.4m \\
{} & Japanese  & 125m & 4.0m \\
{} & Korean & 81.7m & 3.1m \\
{} & Hungarian  & 8.2m & 1.5m \\
{} & Urdu  &  91.5m & 1.0m \\
{} & Amharic & 31m & 15k \\
{} & Azeri  & 24m & 195k \\
{} & Finnish & 5.1m  & 1.4m \\ 
    \bottomrule
    \end{tabular}
    \caption{For AMR and SPARQL parsing, we assemble test sets selecting utterances from two subsets of languages: (1) The presumably easier subset MASSIVE- (top) covering the same Indo-European languages as QALD9, and (2) the more diverse MASSIVE+ (bottom), e.g., targeting different writing systems. Statistics are estimates, based on \footnotesize{\url{https://meta.wikimedia.org/wiki/List\_of\_Wikipedias}} and Google search results.}
    \label{tab:experiment-languages}
\end{table}

\subsection{Evaluation Guidelines}
\noindent
For AMR parsing, we report Smatch \citep{Cai2013}, while for SPARQL we check (1)~query executability (using the Python SPARQLWrapper) and (2)~whether the query returns an answer from DBPedia. We do not check answer factuality, as our objective is to measure model confidence in semantic parse correctness, not the model's knowledge of the contents of a given knowledge base (given that knowledge bases change over time and many local entities do not have a DBPedia entry, for example).

For hallucination detection experiments using in-context learning, we employ quantitative and qualitative means of analysis. For perturbed examples (i.e., parse a query for a question likely to require a known disallowed relation), a predicted ranking is good if the model: (1) ranks the AMR higher, (2) ranks the SPARQL higher yet verifies the relation is not allowed, or (3) produces a valid alternative SPARQL. We stratify results by dataset, check executability and whether the query returns an existing record, and also evaluate manually.

For fine-tuned joint AMR-SPARQL, we use a diverse beam search (n=5) and different methods to determine relative ranking: (1) check the top-ranked produced sequence, (2) count the majority structure in the \texttt{N}-best ranking, and (3) compare transition scores for the first token produced.\footnote{Either `AMR' or `SPARQL,' or the first sub-token therein.}
Our hypothesis is models will prefer SPARQL over AMR for QALD9 and vice versa for MASSIVE-AMR. This is a reasonable hypothesis, as all QALD9 is known to be matched with ground truth SPARQL, while fewer queries in MASSIVE-AMR are likely convertible into an executable query, an assumption we assess qualitatively (\Cref{appendix:long-tail}). 

For evaluation, models output a queryable object (JSON) with three key-value pairs: parsed query, list of relations in query, and list of relation verifications (boolean values) (see \Cref{appendix:prompt}), with very few structural errors observed (<1\% in our studies).

\subsection{Results}
\noindent
We present results on in-context learning for AMR  parsing (\Cref{tab:results-amr-generation}) and SPARQL queries (\Cref{tab:results-sparql-generation}) across languages, report on SPARQL hallucinations (\Cref{tab:results-hallucinations}), followed by results in hallucination detection using in-context joint models (\Cref{tab:results-context-joint}), as well as fine-tuned joint models (\Cref{tab:results-fine-tuned}).

\addtolength{\tabcolsep}{-1pt} 
\begin{table}
    \footnotesize
    \centering
    \begin{tabular}{ l l l  | l l l }
    \toprule
    {} & \small{\textbf{Model}} & \small{\textbf{Data}}  & \small{\textbf{F1}} $\uparrow$ \\
    \midrule
    \parbox[t]{2mm}{\multirow{4}{*}{\rotatebox[origin=c]{90}{\scriptsize{Few-shot/EN}}}} & 
    {GPT-3.5} & {MASSIVE-EN}  & {0.43$_{\pm0.20}$}  \\
    {} & {} & {QALD9-EN}    & {0.57$_{\pm0.17}$}  \\
    {} & {GPT-4} & {MASSIVE-EN}  & {0.53$_{\pm0.21}$} \\
    {} & {} & {QALD9-EN}  & {0.70$_{\pm0.16}$}\\
    \midrule   
    \parbox[t]{2mm}{\multirow{6}{*}{\rotatebox[origin=c]{90}{\scriptsize{Few-shot/non-EN}}}} & 
    {GPT-3.5} & {MASSIVE+}   & {0.33$_{\pm0.22}$}  \\
    {} & {} & {MASSIVE-}   & {0.42$_{\pm0.20}$}  \\
    {} & {} & {QALD9}   & {0.44$_{\pm0.20}$}  \\

    {} & {GPT-4} & {MASSIVE+}   & {0.46$_{\pm0.21}$}  \\
    {} & {} & {MASSIVE-}   & {0.49$_{\pm0.20}$}  \\
    {} & {} & {QALD9}    & {0.58$_{\pm0.22}$}  \\
    \midrule
    \parbox[t]{2mm}{\multirow{2}{*}{\rotatebox[origin=c]{90}{\scriptsize{SOTA}}}} & MBSE & {QALD9-EN}    & {0.90}  \\
    {} & {} & AMR 3.0      & {0.84} \\
    \bottomrule
    \end{tabular}
    \caption{AMR parsing results by model, dataset, and language subset, comparing in-context learning (top and middle) with SOTA \citep{Lee2022} (bottom). Overall, in-context learning is less effective than more engineered approaches.}
    \label{tab:results-amr-generation}
\end{table}

\subsection{Analysis and Discussion}
\noindent
For \textbf{AMR parsing} (Research question \ref{rq1}), results (\Cref{tab:results-amr-generation}, examples and error analysis in \Cref{appendix:amr-generation}) show that state-of-the-art AMR systems still outperform in-context learning with margins between {10-20\%}, a display of the strengths of engineered modular systems, data augmentation, and AMR post-processing. Comparing few-shot models, GPT-4 outperforms GPT-3.5 by a margin of 10-13\% F1, with performance on QALD9 14-17\% F1 higher than MASSIVE-AMR, evidence of the challenge of the latter. Models perform 5-12\% F1 higher for MASSIVE- compared to more diverse MASSIVE+ (see  \Cref{ssec:languages}), the first reported AMR results we are aware of for many of these languages.

\textbf{SPARQL parsing}. Results of SPARQL query parsing with in-context learning (\Cref{tab:results-sparql-generation}, examples in \Cref{appendix:sparql-generation}) provide evidence that LLMs perform well in a few-shot setting, exceeding 90\% F1 in executability across datasets and languages. However, as LLMs are not trained on up-to-date data, no more than 52\% of queries for QALD9 and 32\% of MASSIVE-AMR return existing DBPedia records. Additionally, models display good performance for MASSIVE+, where AMR performance was observed to decrease, evidence that LLMs have more knowledge of SPARQL than AMR structures.

\begin{table}
    \small
    \centering
    \begin{tabular}{ l l | l l }
    \toprule
    {} & {\textbf{Data}}   & {\textbf{Exec.}} $\uparrow$ & {\textbf{Returns}} $\uparrow$ \\

    \midrule
    \parbox[t]{2mm}{\multirow{3}{*}{\rotatebox[origin=c]{90}{{GPT-3.5}}}} & {MASSIVE+} & {0.93}     & {0.32} \\
    {} & {MASSIVE-}     & {0.94}    & {0.41}  \\
    {} & {QALD9}        & {0.97}     & {0.53}  \\
    \midrule
    \parbox[t]{2mm}{\multirow{3}{*}{\rotatebox[origin=c]{90}{{GPT-4}}}} & {MASSIVE+}   & {0.94}     & {0.34}  \\
    {} & {MASSIVE-}    & {0.99}     & {0.50}  \\
    {} & {QALD9}         & {1.00}     & {0.52}  \\
    \bottomrule
    \end{tabular}
    \caption{Few-shot SPARQL parsing results across datasets and models. We report executability and how many return existing records. Overall, models produce structurally viable SPARQL across languages.}
    \label{tab:results-sparql-generation}
\end{table}

\addtolength{\tabcolsep}{-1pt} 
\begin{table}
    \centering
    \footnotesize
    \begin{tabular}{l l l l | l >{\columncolor[gray]{0.8}}l}
    \toprule
    {} & \small{\textbf{Data}} & \small{\textbf{Perturb}} & \small{\textbf{\#Utts}} & \small{\textbf{Halluc.}} $\downarrow$ & \small{\textbf{Detects}} $\uparrow$ \\
    \midrule
    \parbox[t]{1mm}{\multirow{6}{*}{\rotatebox[origin=c]{90}{GPT-3.5}}} & 
    {MASSIVE+} & {No} & {38}  & {0.21} & {0.0} \\

    {} & {} & {\textbf{Yes}} & {62}   & \textbf{0.71} & {0.04} \\
    {} & {MASSIVE-} & {No} & {38}  & {0.16} & {0.0} \\

    {} & {} & {\textbf{Yes}} & {62}  & \textbf{0.59} & {0.0} \\
    
    {} & {QALD9} & {No} & {110}   & {0.22} & {0.09} \\

    {} & {} & {\textbf{Yes}} & {110}   & \textbf{0.84} & {0.0} \\
    \midrule
    \parbox[t]{1mm}{\multirow{6}{*}{\rotatebox[origin=c]{90}{GPT-4}}} & 
    {MASSIVE+} & {No} & {34}   & {0.06}  & {0.50} \\

    {} & {} & \textbf{Yes}  & {66}   & \textbf{0.48} & {0.09} \\
    {} & {MASSIVE-} & {No} & {36}  & {0.0} & {n/a} \\

    {} & {} & \textbf{Yes}  & {64}   & \textbf{0.54} & {0.14} \\
    {} & {QALD9} & {No} & {50}   & {0.04}& {0.0} \\

    {} & {} & \textbf{Yes} & {50}  & \textbf{0.46} & {0.08} \\
    \bottomrule
    \end{tabular}
    \caption{Rates of SPARQL hallucination and hallucination detection with a SPARQL-only model. When we perturb a relation, hallucination is high, that is, models produce top-ranked queries with disallowed relations; in all settings, detection rates (gray) are consistently poor, that is models fail to verify relations are allowed or not.}
    \label{tab:results-hallucinations}
\end{table}

\textbf{SPARQL relation hallucination rates} (Research question \ref{rq2}). In \Cref{tab:results-hallucinations}, we examine if: (1)~models hallucinate SPARQL relations when we remove some relations from an allowed list, and (2)~models also can detect cases of generated relations not being allowed (i.e. hallucinations). In a nutshell, results confirm all models often hallucinate relations and yet fail at detection consistently. 

Specifically, we find that under normal, non-perturbed conditions across languages (odd rows of \Cref{tab:results-hallucinations}), GPT-3.5 exhibits hallucination rates between 16-22\%, which GPT-4 reduces to 0-6\%. When we disallow a relation likely to be needed in the query (rows where Perturb=\textbf{Yes}), hallucination rates increase considerably: for GPT-3.5 to between 40-60\%, and for GPT-4 between 42-54\%.

\textbf{Hallucination detection, non-joint model}. With 2-shot SPARQL query parsing, models show poor rates of hallucination detection (\Cref{tab:results-hallucinations}), with GPT-4 detecting no more than 14\% of all hallucinations. In a vast majority of cases (86-100\%, gray column), models are deceptive, incorrectly reporting that disallowed relations are allowed (Ex. 2 in \Cref{appendix:sparql-generation}), providing us with justification to test if we can do better with a joint AMR-SPARQL model.

\textbf{Hallucination detection, in-context joint model} (Research question \ref{rq3}). Overall, in-context learning for hallucination detection is quite challenging. With oracle knowledge of which relation has been disallowed (\Cref{tab:results-context-joint}), GPT-4 still misreports 24\% of cases. 

Nevertheless, we find evidence that GPT-4 with an oracle employs dual hallucination detection strategies in some cases: for 1 in 5 hallucinations, the model ranks AMRs higher, and, for 3 of 5, it parses queries with disallowed relations which it accurately verifies as non-existent.

Without an oracle, the rate of \textit{deception} (i.e. \textit{not} detecting a hallucinated relation) exceeds 80\% in both cases tested, which proved challenging to overcome despite multiple prompt variations, including promised rewards for sticking to allowed relations, veiled (and unveiled) threats, repeated warnings, and legalese which bound the model to abide by restrictions, tactics the models consistently disregarded, suggesting space for future research into LLM confidence measures for QA as well as structural integrity metrics for a semantic critic.

Considering cases of ambiguous utterances (\textit{hard hallucination detection}), GPT-4 mostly follows the rules (e.g., perhaps parsing `creator' when disallowed for `who created iron man' but verifying correctly the relation is fallacious). However, it is difficult in many cases to qualitatively determine query plausibility for various other relations parsed, as the correctness of any of a large range of queries that models actually produce depends on the target knowledge base, left implicit in our experiments.

\begin{table}
     \small
     \centering
     \begin{tabular}{l l l |  l >{\columncolor[gray]{0.8}}l}
     \toprule
     \small{\textbf{Model}} & \small{\textbf{Oracle}} & \small{\textbf{\#Perturb}} & \small{\textbf{Halluc.}} $\downarrow$ & \small{\textbf{Detects}} $\uparrow$ \\    
     \midrule
       GPT-3.5   & {no}   & {60/120}  & {0.58} & {0.07} \\
     GPT-4    & {no} & {60/120}  & {0.39} & {0.17} \\
     GPT-4  & {yes}  & {150/240}  & {0.31} & {0.76}  \\
     \bottomrule
     \end{tabular}
     \caption{Results of joint AMR-SPARQL detection with in-context learning (8-shot, GPTs), targeting 140 SPARQL relations and 8 languages. Hallucination occurs in at least 1 in 3 cases, and hallucination detection is not effective, except with an oracle (last row).}
     \label{tab:results-context-joint}
\end{table}

\begin{table}[t]
    \small
    \centering
    \begin{tabular}{ l l l | l l l l }
    \toprule
    {} & \textbf{Langs.} & \textbf{Data} &  \small{\textbf{Top1}}  & \small{\textbf{Top5}} & \small{\textbf{Token1}}\\
    \midrule

    \parbox[t]{1mm}{\multirow{4}{*}{\rotatebox[origin=c]{90}{GPT2$_{\textsc{distill}}$}}} 
    & EN & {QALD9} & {0.50 $\newcrossmark$}  &  {0.68 $\checkmark$} & {0.83 $\checkmark$} \\
     & {} & {MASSIVE-AMR} & {0.58}  &  {0.62} & {0.80} \\
     & {Non-EN} & {QALD9} & {0.53 $\newcrossmark$}  &  {0.55 $\newcrossmark$} & {0.74 $\checkmark$} \\
     & {} & {MASSIVE-AMR} & {0.54}  &  {0.54} & {0.70} \\
    \midrule
    \midrule
    \parbox[t]{1mm}{\multirow{4}{*}{\rotatebox[origin=c]{90}{\tiny{LLaMa-13B}}}}  & EN & {QALD9}  & {0.82 $\checkmark$}  &  {0.95 $\newcrossmark$} & {0.90 $\sim$} \\
     & {} & {MASSIVE-AMR} & {0.76}  &  {0.95} & {0.88} \\
     & {Non-EN} & {QALD9} & {0.78 $\newcrossmark$}  &  {0.95 $\newcrossmark$} & {0.82 $\newcrossmark$} \\
     & {} & {MASSIVE-AMR} & {0.88}  &  {0.98} & {0.95} \\
    
    \bottomrule
    \end{tabular}
    \caption{The proportion of cases models prefer SPARQL over AMR structures for QALD9 and MASSIVE-AMR, comparing fine-tuned GPT2-xl$_{\textsc{distill}}$ (top) and Llama-13B (bottom) with English (EN) and non-English data. The hypothesis in each case is that models will prefer SPARQL for QALD9, with a ($\checkmark$) indicating evidence in support. Results from preliminary studies are overall inconclusive.}
    \label{tab:results-fine-tuned}
\end{table}

\textbf{Hallucination detection, fine-tuned joint models} (Research question \ref{rq4}). Results of fine-tuned models are inconclusive (\Cref{tab:results-fine-tuned}). With GPT-2-XL$_{\textrm{distill}}$, preference between SPARQL vs AMR is mostly 50-50, with variation only observed with first token transition scores. LLaMa, in contrast, shows bias towards SPARQL under every condition (between 75-95\%), and only in one setting (top-1) favoring SPARQL consistently for QALD9. Qualitative analysis shows LLaMa prefers AMR for incomplete utterances such as `describe' and `calculate this', yet it often misclassifies currency conversion utterances as having valid SPARQL.\footnote{In principle, currency conversion values could be stored in a knowledge base, but in practice knowledge bases are not updated in real-time.}

With our fine-tuned models, we examined an \texttt{N}-best space from multiple perspectives (top-1 prediction, majority, transition scores). We speculate that the proportion of AMRs versus SPARQL in fine-tuning likely has an effect: in our experiments, we include more AMRs than SPARQL (Appendix \ref{appendix:models}), suggesting a study with  varied proportions of training data is warranted as well as training with more data (we used <6k examples in fine-tuning).

\section{Conclusion}
\noindent
We present MASSIVE-AMR, the largest and most diverse dataset of multilingual questions paired with Abstract Meaning Representation (AMR) graphs, which we publicly release for research purposes. We discuss the origins of the data, and detail the processes of dataset creation, curation, and quality control. 

To examine the utility of our dataset in controlled experimentation using large language models, we first consider the task of \textbf{structure parsing}, showing results for both AMR graph and SPARQL query parsing across languages. Overall, performance for AMR parsing with in-context learning is less effective compared with reported state-of-the-art using fine-tuning; still, qualitative assessment of produced structures reveals many coherent, correct graphs despite low similarity with a ground truth. In comparison, SPARQL parsing performance is high across languages, at least in small studies using the QALD9-AMR dataset.

One motivating factor behind the creation of MASSIVE-AMR was to be able to test the utility of AMRs for knowledge base question answering (KBQA), specifically ascertaining whether AMRs can help \textbf{detect incongruous SPARQL queries}, essentially serving as a proxy confidence measure for the correctness of an answer suggested by a QA system.
In these experiments, we first confirm that the GPT models do indeed hallucinate semantic relations, and then discover that `easy' hallucination detection---asking a model to verify relations are allowed---is actually quite challenging, even for GPT-4. Further, `hard' hallucination detection---the identification of utterances that are likely ambiguous---is \emph{also} challenging, with a joint AMR-SPARQL model only detecting 1 in 5 cases. 

 Beyond the AMR-for-KBQA investigations we performed in this work, we hope that the release of MASSIVE-AMR will support additional research into using structured meaning representations for multilingual QA and model interpretability.

\section{Acknowledgements}
We would like to thank Jack FitzGerald for help with the data and comments on early drafts, Nathan Schneider for feedback on writing and overall project clarity, and Meagan Vigus for thinking about data collection and discussing typological questions with us. We are grateful to Amazon Cambridge linguists and engineers Isabel Groves, Elizabeth Schett, Daraksha Parveen, and Wei-Wei Cheng for help building the dataset and compiling annotation guidelines. Our special thanks to the anonymous reviewers for insightful questions and critique helping us to achieve a more reader-friendly and scientific presentation of ideas. This work was supported through funding from Amazon and DARPA I2O (RA-21-02).

\section{Ethical Considerations}
\textbf{Informed Consent}: We ensured that all individuals providing annotations were fully informed about the purpose of the annotation task, how their data will be used, and what rights they have in relation to their data.

\textbf{Fair Compensation}: We ensured that individuals providing annotations were fairly compensated for their time and effort. For this project, professional annotators were compensated at least \$30/hour, working between 20-80 hours each for the duration of data collection.

\textbf{Transparency}: We were transparent about the purpose and scope of the annotation task, as well as the potential benefits of the project, helping to build trust with individuals providing annotations and ensuring that they understood the significance of their contributions. We intend that through these practices data annotation efforts are overall more effective, resulting in a higher quality resource.

\textbf{Environmental impact}: We considered the environmental impact of the research, including the energy consumption of  computing resources used. With GPT-4 inference, we limited input to 100s of examples to reduce costs. In-house fine-tuning was done using parameter efficient fine-tuning methods, allowing each experiment to be done on 1-2 NVIDIA Quadro RTX 8000 GPUs in <24 hours.

\section{Limitations}

\begin{enumerate}
\item Our work involved research into multilingual SPARQL and AMR parsing; though our dataset includes 52 languages, we report results on no more than 10-12 of these. Many of the languages we included are Indo-European, with only a few exceptions (Korean, Japanese, Amharic, Vietnamese). 
\item No experiments in joint AMR-SPARQL parsing involved hypotheses about performance across languages, though some evidence of performance shifts has been observed.
\item Fine-tuning models was done with less than 6k AMRs and 3-4k SPARQL training examples. Test data was limited to 100s examples per language in order to allow for multiple iterations and explore hyperparameter settings. Increasing the sizes of training and test sets is left for future work.
\item Testing was limited to four large language models in this work (GPT-2-XL$_{\text{distill}}$, GPT-3.5, GPT-4, LLaMa). LLaMa does include multilingual data in training \citep{Touvron2023llama}, particularly languages using Latin and Cyrillic scripts. We did not test models explicitly trained for multilingual purposes and for other scripts, leaving such work for the future.
\item The MASSIVE-AMR dataset matches multilingual utterances to unique AMR graphs, making it the largest such dataset to date. However, unlike QALD9-AMR \citep{Lee2022}, MASSIVE-AMR does not include gold SPARQL queries. We emphasize that the use case we explore in this paper is only one of many possible, and we hope future research explores beyond this single application.
\end{enumerate}

\section{Bibliographical References}\label{sec:reference}
\bibliography{anthology,custom}

\begin{thebibliography}{46}
\expandafter\ifx\csname natexlab\endcsname\relax\def\natexlab#1{#1}\fi

\bibitem[{Aksitov et~al.(2023)Aksitov, Chang, Reitter, Shakeri, and Sung}]{Aksitov2023}
Renat Aksitov, Chung-Ching Chang, David Reitter, Siamak Shakeri, and Yunhsuan Sung. 2023.
\newblock \href {http://arxiv.org/abs/2302.05578} {Characterizing attribution and fluency tradeoffs for retrieval-augmented large language models}.
\newblock \emph{arXiv:2302.05578}.

\bibitem[{Banarescu et~al.(2013)Banarescu, Bonial, Cai, Georgescu, Griffitt, Hermjakob, Knight, Koehn, Palmer, and Schneider}]{Banarescu2013}
Laura Banarescu, Claire Bonial, Shu Cai, Madalina Georgescu, Kira Griffitt, Ulf Hermjakob, Kevin Knight, Philipp Koehn, Martha Palmer, and Nathan Schneider. 2013.
\newblock \href {https://aclanthology.org/W13-2322} {{A}bstract {M}eaning {R}epresentation for sembanking}.
\newblock In \emph{Proceedings of the 7th Linguistic Annotation Workshop and Interoperability with Discourse}, pages 178--186, Sofia, Bulgaria. Association for Computational Linguistics.

\bibitem[{Banarescu et~al.(2019)Banarescu, Bonial, Cai, Georgescu, Griffitt, Hermjakob, Knight, Koehn, Palmer, and Schneider}]{AMRGuidelines}
Laura Banarescu, Claire Bonial, Shu Cai, Madalina Georgescu, Kira Griffitt, Ulf Hermjakob, Kevin Knight, Philipp Koehn, Martha Palmer, and Nathan Schneider. 2019.
\newblock Abstract {M}eaning {R}epresentation ({AMR}) 1.2.6 specification.
\newblock \url{https://github.com/amrisi/amr-guidelines/blob/master/amr.md}.

\bibitem[{Banerjee et~al.(2022)Banerjee, Nair, Kaur, Usbeck, and Biemann}]{Banerjee2022}
Debayan Banerjee, Pranav~Ajit Nair, Jivat~Neet Kaur, Ricardo Usbeck, and Chris Biemann. 2022.
\newblock \href {https://doi.org/10.1145%2F3477495.3531841} {Modern baselines for {SPARQL} semantic parsing}.
\newblock In \emph{Proceedings of the 45th International {ACM} {SIGIR} Conference on Research and Development in Information Retrieval}. {ACM}.

\bibitem[{Bevilacqua et~al.(2021)Bevilacqua, Blloshmi, and Navigli}]{Bevilacqua2021}
Michele Bevilacqua, Rexhina Blloshmi, and Roberto Navigli. 2021.
\newblock One {SPRING} to rule them both: {S}ymmetric {AMR} semantic parsing and generation without a complex pipeline.
\newblock In \emph{Proceedings of AAAI}.

\bibitem[{Bornea et~al.(2022)Bornea, Astudillo, Naseem, Mihindukulasooriya, Abdelaziz, Kapanipathi, Florian, and Roukos}]{Bornea2022}
Mihaela Bornea, Ramon~Fernandez Astudillo, Tahira Naseem, Nandana Mihindukulasooriya, Ibrahim Abdelaziz, Pavan Kapanipathi, Radu Florian, and Salim Roukos. 2022.
\newblock \href {http://arxiv.org/abs/2112.07877} {Learning to transpile {AMR} into {SPARQL}}.
\newblock \emph{arXiv:2112.07877}.

\bibitem[{Cai and Knight(2013)}]{Cai2013}
Shu Cai and Kevin Knight. 2013.
\newblock \href {https://aclanthology.org/P13-2131} {{S}match: an evaluation metric for semantic feature structures}.
\newblock In \emph{Proceedings of the 51st Annual Meeting of the Association for Computational Linguistics}, pages 748--752, Sofia, Bulgaria. Association for Computational Linguistics.

\bibitem[{Croft(2002)}]{Croft2002}
William Croft. 2002.
\newblock \emph{Typology and Universals}.
\newblock Cambridge Textbooks in Linguistics. Cambridge University Press.

\bibitem[{Cui et~al.(2022)Cui, Aralikatte, Lent, and Hershcovich}]{Cui2022}
Ruixiang Cui, Rahul Aralikatte, Heather Lent, and Daniel Hershcovich. 2022.
\newblock \href {https://doi.org/10.1162/tacl_a_00499} {{Compositional Generalization in Multilingual Semantic Parsing over Wikidata}}.
\newblock \emph{Transactions of the Association for Computational Linguistics}, 10:937--955.

\bibitem[{Das et~al.(2018)Das, Munkhdalai, Yuan, Trischler, and McCallum}]{Das2018}
Rajarshi Das, Tsendsuren Munkhdalai, Xingdi Yuan, Adam Trischler, and Andrew McCallum. 2018.
\newblock \href {http://arxiv.org/abs/1810.05682} {Building dynamic knowledge graphs from text using machine reading comprehension}.
\newblock \emph{arXiv:1810.05682}.

\bibitem[{Drozdov et~al.(2022)Drozdov, Zhou, Florian, McCallum, Naseem, Kim, and Astudillo}]{Drozdov2022}
Andrew Drozdov, Jiawei Zhou, Radu Florian, Andrew McCallum, Tahira Naseem, Yoon Kim, and Ram{\'o}n Astudillo. 2022.
\newblock \href {https://doi.org/10.18653/v1/2022.naacl-main.80} {Inducing and using alignments for transition-based {AMR} parsing}.
\newblock In \emph{Proceedings of the 2022 Conference of the North American Chapter of the Association for Computational Linguistics: Human Language Technologies}, pages 1086--1098, Seattle, United States. Association for Computational Linguistics.

\bibitem[{Ettinger et~al.(2023)Ettinger, Hwang, Pyatkin, Bhagavatula, and Choi}]{Ettinger2023}
Allyson Ettinger, Jena Hwang, Valentina Pyatkin, Chandra Bhagavatula, and Yejin Choi. 2023.
\newblock \href {https://doi.org/10.18653/v1/2023.findings-emnlp.553} {{``}{Y}ou are an expert linguistic annotator{''}: Limits of {LLM}s as analyzers of {A}bstract {M}eaning {R}epresentation}.
\newblock In \emph{Findings of the Association for Computational Linguistics: EMNLP 2023}, pages 8250--8263, Singapore. Association for Computational Linguistics.

\bibitem[{Feng et~al.(2022)Feng, Yang, Cer, Arivazhagan, and Wang}]{feng2020language}
Fangxiaoyu Feng, Yinfei Yang, Daniel Cer, Naveen Arivazhagan, and Wei Wang. 2022.
\newblock \href {https://doi.org/10.18653/v1/2022.acl-long.62} {Language-agnostic {BERT} {S}entence {E}mbedding}.
\newblock In \emph{Proceedings of the 60th Annual Meeting of the Association for Computational Linguistics (Volume 1: Long Papers)}, pages 878--891, Dublin, Ireland. Association for Computational Linguistics.

\bibitem[{FitzGerald et~al.(2023)FitzGerald, Hench, Peris, Mackie, Rottmann, Sanchez, Nash, Urbach, Kakarala, Singh, Ranganath, Crist, Britan, Leeuwis, Tur, and Natarajan}]{Fitzgerald2023}
Jack FitzGerald, Christopher Hench, Charith Peris, Scott Mackie, Kay Rottmann, Ana Sanchez, Aaron Nash, Liam Urbach, Vishesh Kakarala, Richa Singh, Swetha Ranganath, Laurie Crist, Misha Britan, Wouter Leeuwis, Gokhan Tur, and Prem Natarajan. 2023.
\newblock \href {https://doi.org/10.18653/v1/2023.acl-long.235} {{MASSIVE}: A 1{M}-example multilingual natural language understanding dataset with 51 typologically-diverse languages}.
\newblock In \emph{Proceedings of the 61st Annual Meeting of the Association for Computational Linguistics (Volume 1: Long Papers)}, pages 4277--4302, Toronto, Canada. Association for Computational Linguistics.

\bibitem[{Flanigan et~al.(2014)Flanigan, Thomson, Carbonell, Dyer, and Smith}]{Flanigan2014}
Jeffrey Flanigan, Sam Thomson, Jaime Carbonell, Chris Dyer, and Noah~A. Smith. 2014.
\newblock \href {https://doi.org/10.3115/v1/P14-1134} {A discriminative graph-based parser for the {A}bstract {M}eaning {R}epresentation}.
\newblock In \emph{Proceedings of the 52nd Annual Meeting of the Association for Computational Linguistics (Volume 1: Long Papers)}, pages 1426--1436, Baltimore, Maryland. Association for Computational Linguistics.

\bibitem[{Gabriel et~al.(2021)Gabriel, Celikyilmaz, Jha, Choi, and Gao}]{Gabriel2021}
Saadia Gabriel, Asli Celikyilmaz, Rahul Jha, Yejin Choi, and Jianfeng Gao. 2021.
\newblock \href {https://doi.org/10.18653/v1/2021.findings-acl.42} {{GO} {FIGURE}: A meta evaluation of factuality in summarization}.
\newblock In \emph{Findings of the Association for Computational Linguistics: ACL-IJCNLP 2021}, pages 478--487, Online. Association for Computational Linguistics.

\bibitem[{Haji{\v{c}} et~al.(2014)Haji{\v{c}}, Bojar, and Ure{\v{s}}ov{\'a}}]{hajic-etal-2014-comparing}
Jan Haji{\v{c}}, Ond{\v{r}}ej Bojar, and Zde{\v{n}}ka Ure{\v{s}}ov{\'a}. 2014.
\newblock \href {https://doi.org/10.3115/v1/W14-5808} {Comparing {C}zech and {E}nglish {AMR}s}.
\newblock In \emph{Proceedings of Workshop on Lexical and Grammatical Resources for Language Processing}, pages 55--64, Dublin, Ireland. Association for Computational Linguistics and Dublin City University.

\bibitem[{Ji et~al.(2023)Ji, Lee, Frieske, Yu, Su, Xu, Ishii, Bang, Madotto, and Fung}]{Ji2023}
Ziwei Ji, Nayeon Lee, Rita Frieske, Tiezheng Yu, Dan Su, Yan Xu, Etsuko Ishii, Ye~Jin Bang, Andrea Madotto, and Pascale Fung. 2023.
\newblock \href {https://doi.org/10.1145/3571730} {Survey of hallucination in natural language generation}.
\newblock \emph{ACM Comput. Surv.}, 55(12).

\bibitem[{Jiang et~al.(2021)Jiang, Araki, Ding, and Neubig}]{Jiang2021}
Zhengbao Jiang, Jun Araki, Haibo Ding, and Graham Neubig. 2021.
\newblock \href {https://doi.org/10.1162/tacl_a_00407} {{How Can We Know When Language Models Know? On the Calibration of Language Models for Question Answering}}.
\newblock \emph{Transactions of the Association for Computational Linguistics}, 9:962--977.

\bibitem[{Kadavath et~al.(2022)Kadavath, Conerly, Askell, Henighan, Drain, Perez, Schiefer, Hatfield-Dodds, DasSarma, Tran-Johnson, Johnston, El-Showk, Jones, Elhage, Hume, Chen, Bai, Bowman, Fort, Ganguli, Hernandez, Jacobson, Kernion, Kravec, Lovitt, Ndousse, Olsson, Ringer, Amodei, Brown, Clark, Joseph, Mann, McCandlish, Olah, and Kaplan}]{Kadavath2022}
Saurav Kadavath, Tom Conerly, Amanda Askell, Tom Henighan, Dawn Drain, Ethan Perez, Nicholas Schiefer, Zac Hatfield-Dodds, Nova DasSarma, Eli Tran-Johnson, Scott Johnston, Sheer El-Showk, Andy Jones, Nelson Elhage, Tristan Hume, Anna Chen, Yuntao Bai, Sam Bowman, Stanislav Fort, Deep Ganguli, Danny Hernandez, Josh Jacobson, Jackson Kernion, Shauna Kravec, Liane Lovitt, Kamal Ndousse, Catherine Olsson, Sam Ringer, Dario Amodei, Tom Brown, Jack Clark, Nicholas Joseph, Ben Mann, Sam McCandlish, Chris Olah, and Jared Kaplan. 2022.
\newblock \href {https://arxiv.org/abs/2207.05221} {Language models (mostly) know what they know}.
\newblock \emph{arXiv}.

\bibitem[{Kapanipathi et~al.(2021)Kapanipathi, Abdelaziz, Ravishankar, Roukos, Gray, Fernandez~Astudillo, Chang, Cornelio, Dana, Fokoue, Garg, Gliozzo, Gurajada, Karanam, Khan, Khandelwal, Lee, Li, Luus, Makondo, Mihindukulasooriya, Naseem, Neelam, Popa, Gangi~Reddy, Riegel, Rossiello, Sharma, Bhargav, and Yu}]{Kapanipathi2021}
Pavan Kapanipathi, Ibrahim Abdelaziz, Srinivas Ravishankar, Salim Roukos, Alexander Gray, Ram{\'o}n Fernandez~Astudillo, Maria Chang, Cristina Cornelio, Saswati Dana, Achille Fokoue, Dinesh Garg, Alfio Gliozzo, Sairam Gurajada, Hima Karanam, Naweed Khan, Dinesh Khandelwal, Young-Suk Lee, Yunyao Li, Francois Luus, Ndivhuwo Makondo, Nandana Mihindukulasooriya, Tahira Naseem, Sumit Neelam, Lucian Popa, Revanth Gangi~Reddy, Ryan Riegel, Gaetano Rossiello, Udit Sharma, G~P~Shrivatsa Bhargav, and Mo~Yu. 2021.
\newblock \href {https://aclanthology.org/2021.findings-acl.339} {Leveraging {A}bstract {M}eaning {R}epresentation for knowledge base question answering}.
\newblock In \emph{Findings of the Association for Computational Linguistics: ACL-IJCNLP 2021}, pages 3884--3894, Online. Association for Computational Linguistics.

\bibitem[{Lee et~al.(2022)Lee, Astudillo, Thanh~Lam, Naseem, Florian, and Roukos}]{Lee2022}
Young-Suk Lee, Ram{\'o}n Astudillo, Hoang Thanh~Lam, Tahira Naseem, Radu Florian, and Salim Roukos. 2022.
\newblock \href {https://aclanthology.org/2022.naacl-main.393} {Maximum {B}ayes {S}match ensemble distillation for {AMR} parsing}.
\newblock In \emph{Proceedings of the 2022 Conference of the North American Chapter of the Association for Computational Linguistics: Human Language Technologies}, pages 5379--5392, Seattle, United States. ACL.

\bibitem[{Lehmann et~al.(2015)Lehmann, Isele, Jakob, Jentzsch, Kontokostas, Mendes, Hellmann, Morsey, van Kleef, Auer, and Bizer}]{Lehmann2015}
Jens Lehmann, Robert Isele, Max Jakob, Anja Jentzsch, Dimitris Kontokostas, Pablo~N. Mendes, Sebastian Hellmann, Mohamed Morsey, Patrick van Kleef, S{\"o}ren Auer, and Christian Bizer. 2015.
\newblock \href {http://jens-lehmann.org/files/2015/swj_dbpedia.pdf} {{DBpedia} - a large-scale, multilingual knowledge base extracted from wikipedia}.
\newblock \emph{Semantic Web Journal}, 6(2):167--195.

\bibitem[{Lewis et~al.(2020)Lewis, Perez, Piktus, Petroni, Karpukhin, Goyal, K\"{u}ttler, Lewis, Yih, Rockt\"{a}schel, Riedel, and Kiela}]{Lewis2020}
Patrick Lewis, Ethan Perez, Aleksandra Piktus, Fabio Petroni, Vladimir Karpukhin, Naman Goyal, Heinrich K\"{u}ttler, Mike Lewis, Wen-tau Yih, Tim Rockt\"{a}schel, Sebastian Riedel, and Douwe Kiela. 2020.
\newblock \href {https://proceedings.neurips.cc/paper_files/paper/2020/file/6b493230205f780e1bc26945df7481e5-Paper.pdf} {Retrieval-augmented generation for knowledge-intensive nlp tasks}.
\newblock In \emph{Advances in Neural Information Processing Systems}, volume~33, pages 9459--9474. Curran Associates, Inc.

\bibitem[{Lewis et~al.(2021)Lewis, Stenetorp, and Riedel}]{Lewis2021}
Patrick Lewis, Pontus Stenetorp, and Sebastian Riedel. 2021.
\newblock \href {https://doi.org/10.18653/v1/2021.eacl-main.86} {Question and answer test-train overlap in open-domain question answering datasets}.
\newblock In \emph{Proceedings of the 16th Conference of the European Chapter of the Association for Computational Linguistics: Main Volume}, pages 1000--1008, Online. Association for Computational Linguistics.

\bibitem[{Navigli et~al.(2022)Navigli, Blloshmi, and Martinez~Lorenzo}]{Navigli2022}
Roberto Navigli, Rexhina Blloshmi, and Abelardo Martinez~Lorenzo. 2022.
\newblock {B}abelnet {M}eaning {R}epresentation: A fully semantic formalism to overcome language barriers.
\newblock In \emph{Proceedings of the AAAI Conference on Artificial Intelligence}.

\bibitem[{Niu et~al.(2023)Niu, Huang, Liu, Cui, Wang, and Huang}]{niu2023bridging}
Yilin Niu, Fei Huang, Wei Liu, Jianwei Cui, Bin Wang, and Minlie Huang. 2023.
\newblock \href {https://doi.org/10.1162/tacl_a_00552} {Bridging the gap between synthetic and natural questions via sentence decomposition for semantic parsing}.
\newblock \emph{Transactions of the Association for Computational Linguistics}, 11:367--383.

\bibitem[{OpenAI(2023)}]{openai2023GPT4}
OpenAI. 2023.
\newblock \href {http://arxiv.org/abs/2303.08774} {Gpt-4 technical report}.
\newblock \emph{arXiv:2303.08774}.

\bibitem[{Opitz(2023)}]{opitz-2023-smatch}
Juri Opitz. 2023.
\newblock \href {https://aclanthology.org/2023.findings-eacl.118} {{SMATCH}++: Standardized and extended evaluation of semantic graphs}.
\newblock In \emph{Findings of the Association for Computational Linguistics: EACL 2023}, pages 1595--1607, Dubrovnik, Croatia. Association for Computational Linguistics.

\bibitem[{Opitz et~al.(2020)Opitz, Parcalabescu, and Frank}]{s2match/tacl_a_00329}
Juri Opitz, Letitia Parcalabescu, and Anette Frank. 2020.
\newblock \href {https://doi.org/10.1162/tacl_a_00329} {{AMR} similarity metrics from principles}.
\newblock \emph{Transactions of the Association for Computational Linguistics}, 8:522--538.

\bibitem[{Perevalov et~al.(2022)Perevalov, Diefenbach, Usbeck, and Both}]{Perevalov2022}
Aleksandr Perevalov, Dennis Diefenbach, Ricardo Usbeck, and Andreas Both. 2022.
\newblock Qald-9-plus: A multilingual dataset for question answering over dbpedia and wikidata translated by native speakers.
\newblock In \emph{2022 IEEE 16th International Conference on Semantic Computing (ICSC)}.

\bibitem[{Qiu et~al.(2023)Qiu, Ziser, Korhonen, Ponti, and Cohen}]{Qiu2023}
Yifu Qiu, Yftah Ziser, Anna Korhonen, Edoardo Ponti, and Shay Cohen. 2023.
\newblock \href {https://doi.org/10.18653/v1/2023.emnlp-main.551} {Detecting and mitigating hallucinations in multilingual summarisation}.
\newblock In \emph{Proceedings of the 2023 Conference on Empirical Methods in Natural Language Processing}, pages 8914--8932, Singapore. Association for Computational Linguistics.

\bibitem[{Rashkin et~al.(2023)Rashkin, Nikolaev, Lamm, Aroyo, Collins, Das, Petrov, Tomar, Turc, and Reitter}]{Rashkin2023}
Hannah Rashkin, Vitaly Nikolaev, Matthew Lamm, Lora Aroyo, Michael Collins, Dipanjan Das, Slav Petrov, Gaurav~Singh Tomar, Iulia Turc, and David Reitter. 2023.
\newblock \href {https://doi.org/10.1162/coli_a_00486} {{Measuring Attribution in Natural Language Generation Models}}.
\newblock \emph{Computational Linguistics}, pages 1--64.

\bibitem[{Talmor and Berant(2018)}]{Talmor2018}
Alon Talmor and Jonathan Berant. 2018.
\newblock \href {https://doi.org/10.18653/v1/N18-1059} {The web as a knowledge-base for answering complex questions}.
\newblock In \emph{Proceedings of the 2018 Conference of the North {A}merican Chapter of the Association for Computational Linguistics: Human Language Technologies, Volume 1 (Long Papers)}, pages 641--651, New Orleans, Louisiana. Association for Computational Linguistics.

\bibitem[{Touvron et~al.(2023)Touvron, Lavril, Izacard, Martinet, Lachaux, Lacroix, Rozière, Goyal, Hambro, Azhar, Rodriguez, Joulin, Grave, and Lample}]{Touvron2023llama}
Hugo Touvron, Thibaut Lavril, Gautier Izacard, Xavier Martinet, Marie-Anne Lachaux, Timothée Lacroix, Baptiste Rozière, Naman Goyal, Eric Hambro, Faisal Azhar, Aurelien Rodriguez, Armand Joulin, Edouard Grave, and Guillaume Lample. 2023.
\newblock \href {http://arxiv.org/abs/2302.13971} {Llama: Open and efficient foundation language models}.
\newblock \emph{arXiv:2302.13971}.

\bibitem[{Usbeck et~al.(2018)Usbeck, Gusmita, Ngomo, and Saleem}]{Usbeck2018}
Ricardo Usbeck, Ria Gusmita, Axel-Cyrille~Ngonga Ngomo, and Muhammad Saleem. 2018.
\newblock 9th challenge on question answering over linked data (qald-9).
\newblock In \emph{Joint proceedings of the 4th Workshop on Semantic Deep Learning (SemDeep-4) and NLIWoD4}.

\bibitem[{Van~Gysel et~al.(2021)Van~Gysel, Vigus, Chun, Lai, Moeller, Yao, O’Gorman, Cowell, Croft, Huang, Hajič, Martin, Oepen, Palmer, Pustejovsky, Vallejos, and Xue}]{Gysel2021}
Jens Van~Gysel, Meagan Vigus, Jayeol Chun, Kenneth Lai, Sarah Moeller, Jiarui Yao, Tim O’Gorman, Andrew Cowell, William Croft, Chu-Ren Huang, Jan Hajič, James Martin, Stephan Oepen, Martha Palmer, James Pustejovsky, Rosa Vallejos, and Nianwen Xue. 2021.
\newblock Designing a uniform meaning representation for natural language processing.
\newblock \emph{Künstliche Intelligenz}.

\bibitem[{Wang et~al.(2023)Wang, Xu, Guo, Hu, Bai, Zhang, and Zhang}]{wang2023exploiting}
Cunxiang Wang, Zhikun Xu, Qipeng Guo, Xiangkun Hu, Xuefeng Bai, Zheng Zhang, and Yue Zhang. 2023.
\newblock \href {https://doi.org/10.18653/v1/2023.findings-acl.131} {Exploiting {A}bstract {M}eaning {R}epresentation for open-domain question answering}.
\newblock In \emph{Findings of the Association for Computational Linguistics: ACL 2023}, pages 2083--2096, Toronto, Canada. Association for Computational Linguistics.

\bibitem[{Wein et~al.(2022)Wein, Leung, Mu, and Schneider}]{wein-etal-2022-effect}
Shira Wein, Wai~Ching Leung, Yifu Mu, and Nathan Schneider. 2022.
\newblock \href {https://aclanthology.org/2022.law-1.12} {Effect of source language on {AMR} structure}.
\newblock In \emph{Proceedings of the 16th Linguistic Annotation Workshop (LAW-XVI) within LREC2022}, pages 97--102, Marseille, France. European Language Resources Association.

\bibitem[{Wein and Schneider(2022)}]{Wein2022}
Shira Wein and Nathan Schneider. 2022.
\newblock \href {https://aclanthology.org/2022.coling-1.336} {Accounting for language effect in the evaluation of cross-lingual {AMR} parsers}.
\newblock In \emph{Proceedings of the 29th International Conference on Computational Linguistics}, pages 3824--3834, Gyeongju, Republic of Korea. International Committee on Computational Linguistics.

\bibitem[{Wein and Schneider(2024)}]{10.1162/coli_a_00503}
Shira Wein and Nathan Schneider. 2024.
\newblock \href {https://doi.org/10.1162/coli_a_00503} {{Assessing the Cross-linguistic Utility of Abstract Meaning Representation}}.
\newblock \emph{Computational Linguistics}, pages 1--55.

\bibitem[{West et~al.(2022)West, Bhagavatula, Hessel, Hwang, Jiang, Le~Bras, Lu, Welleck, and Choi}]{West2022}
Peter West, Chandra Bhagavatula, Jack Hessel, Jena Hwang, Liwei Jiang, Ronan Le~Bras, Ximing Lu, Sean Welleck, and Yejin Choi. 2022.
\newblock \href {https://doi.org/10.18653/v1/2022.naacl-main.341} {Symbolic knowledge distillation: from general language models to commonsense models}.
\newblock In \emph{Proceedings of the 2022 Conference of the North American Chapter of the Association for Computational Linguistics: Human Language Technologies}, pages 4602--4625, Seattle, United States. Association for Computational Linguistics.

\bibitem[{Xue et~al.(2014)Xue, Bojar, Haji{\v{c}}, Palmer, Ure{\v{s}}ov{\'a}, and Zhang}]{Xue2014}
Nianwen Xue, Ond{\v{r}}ej Bojar, Jan Haji{\v{c}}, Martha Palmer, Zde{\v{n}}ka Ure{\v{s}}ov{\'a}, and Xiuhong Zhang. 2014.
\newblock \href {http://www.lrec-conf.org/proceedings/lrec2014/pdf/384_Paper.pdf} {Not an interlingua, but close: Comparison of {E}nglish {AMR}s to {C}hinese and {C}zech}.
\newblock In \emph{Proceedings of the Ninth International Conference on Language Resources and Evaluation ({LREC}'14)}, pages 1765--1772, Reykjavik. European Language Resources Association.

\bibitem[{Zelle and Mooney(1996)}]{Zelle1996}
John~M. Zelle and Raymond~J. Mooney. 1996.
\newblock Learning to parse database queries using inductive logic programming.
\newblock In \emph{Proceedings of the Thirteenth National Conference on Artificial Intelligence - Volume 2}, AAAI'96, page 1050–1055. AAAI Press.

\bibitem[{Zettlemoyer and Collins(2005)}]{Zettlemoyer2005}
Luke~S. Zettlemoyer and Michael Collins. 2005.
\newblock Learning to map sentences to logical form: Structured classification with probabilistic categorial grammars.
\newblock In \emph{Proceedings of the Twenty-First Conference on Uncertainty in Artificial Intelligence}, UAI'05, page 658–666, Arlington, Virginia, USA. AUAI Press.

\bibitem[{Zhang et~al.(2023)Zhang, Chen, Xu, Cao, Chen, Cohn, and Fang}]{Zhang2023}
Qin Zhang, Shangsi Chen, Dongkuan Xu, Qingqing Cao, Xiaojun Chen, Trevor Cohn, and Meng Fang. 2023.
\newblock \href {https://doi.org/10.18653/v1/2023.acl-long.808} {A survey for efficient open domain question answering}.
\newblock In \emph{Proceedings of the 61st Annual Meeting of the Association for Computational Linguistics (Volume 1: Long Papers)}, pages 14447--14465, Toronto, Canada. Association for Computational Linguistics.

\end{thebibliography}
\bibliographystyle{acl_natbib}

\clearpage
\section{Appendices}
\appendix

\section{Characterizing Massive-AMR}
\label{appendix:massive}

\subsection{AMR Top Nodes Across Datasets}
\begin{table}[h]
    \centering
    \scriptsize
    \begin{tabular}{rc|rc|rc}
    \toprule
    \textbf{AMR 3.0} & {\#} & \textbf{QALD9-AMR} & {\#} & \textbf{MASSIVE-AMR} & {\#} \\
    \midrule
and & {7k} &  give-01 & {76} & rate-01 & {105} \\
say-01 & {3k} & have-03 & {50} & define-01 & {103} \\
contrast-01 & {3k} & have-degree & {27} & tell-01 & {94} \\
multi-sentence & {1.7k} &  have-org-role & {21} & have-quant & {87} \\
possible-01 & {1.7k} & be-located-at  & {15}  & equal-01 & {86} \\
cause-01 & {1.6k} &  die-01 & {14} & price-01 & {70}\\
state-01 & {1.5k} &  write-01 & {14} & describe-01 & {66} \\
have-concession & {944} &  bear-02 & {13} & be-located-at & {64} \\
think-01 & {901} &  marry-01 & {13} &  person & {58}\\
person & {705} & show-01 & {12} &  mean-01 & {50}\\
have-03 & {618} & locate-01 & {10} & have-degree & {50}\\
have-condition & {605}& have-rel-role & {10} & bear-02 & {46} \\
date-entity & {538} & person & {9} &  have-org-role & {32}\\
know-01 & {451} & name-01 & {9} & show-01 & {21} \\
have-degree & {440} &  list-01 & {8} & find-01 & {21} \\ 
\bottomrule
    \end{tabular}
    \caption{15 most frequent top AMR nodes in AMR 3.0, QALD9-AMR and MASSIVE-AMR, with counts for a single language (English).}
    \label{tab:top-nodes}
\end{table}

\subsection{Describing the MASSIVE Long Tail}
\label{appendix:long-tail}

We note long-tail characteristics of utterances in MASSIVE \citep{Fitzgerald2023}.
\begin{itemize}[itemsep=0.0pt]
    \item Outliers in terms of utterance length: some 1-2 tokens, others quite long (40+ tokens)
    \item Ambiguous referents (`chase' in `is chase doing good' could be a bank, person, or activity)
    \item Incomplete arithmetic (`tell me what equals two three')
    \item Less frequent expressions (`who is the better half of obama')
    \item Incomplete questions (`synonym for word', `is equal to', `research someone')
\end{itemize}

\pagebreak

\section{Model Details}
\label{appendix:models}

For experiments in joint AMR-SPARQL hallucination and hallucination detection, we tested both fine-tuned models (Table \ref{tab:finetuning}) and in-context learning (Table \ref{tab:in-context}).

\begin{table}[h]
    \footnotesize
    \centering
    \begin{tabular}{l l|l}
    \toprule
    {} & \textbf{Element} & \textbf{Detail} \\
    \midrule
    \parbox[t]{1mm}{\multirow{9}{*}{\rotatebox[origin=l]{90}{\textbf{Fine-tuning}}}} &
    Train set (QALD9/MASSIVE-AMR)  & 6000/2000 \\
    {} & Train set (SPARQL/AMR)   & 3000/5000 \\
    {} & Train set (langs)    & 1300 each, 6 lgs \\
    {} & Test set (QALD9/MASSIVE-AMR)   & 700/500 \\
    {} & Test set (langs)    & 200 each, 6 lgs \\
    {} & Block size (GPT-2/LlaMa) & 512/2048 \\
    {} & Number epochs & 8-16 \\
    {} & Learning rate & $3e^{-5}$ \\
    {} & Optimizer & AdamW \\
    \midrule
    \parbox[t]{1mm}{\multirow{6}{*}{\rotatebox[origin=l]{90}{\textbf{Inference}}}} & Number beams & 20 \\
    {} & Beam size & 5 \\
    {} & Number beam groups & 10 \\
    {} & Diversity penalty & 1.0 \\
    {} & Minimum length & 8 \\
    {} & Maximum length & 256 \\
    \bottomrule
    \end{tabular}
    \caption{Details about training and test splits (top), with model parameters for fine-tuning GPT-2-XL$_{\text{distill}}$ and LlaMa using Hugging Face transformers and PEFT.}
    \label{tab:finetuning}
\end{table}

\begin{table}
    \footnotesize
    \centering
    \begin{tabular}{l|l}
    \toprule
    \textbf{Element} & \textbf{Detail} \\
    \midrule
    Number in-context exs.  & 8-12 \\
    Number tokens in prompt & 2400 \\ 
    In-context langs.    & English, Spanish \\
    Test set (QALD9/MASSIVE-AMR)   & 150/150 \\
    Test set (langs)    & 20-40 each, 10 lgs \\
    Temperature & 1.0 \\
    \bottomrule
    \end{tabular}
    \caption{Details about in-context learning (GPT-3.5 and GPT-4-0613).}
    \label{tab:in-context}
\end{table}

\pagebreak

\onecolumn
\section{Example prompts}
\label{appendix:prompt}
\subsection{SPARQL Parsing (Research question 1b)}
\begin{figure*}[ht]
\begin{lstlisting}[basicstyle=\scriptsize]
{["messages": [
    {
        "role": "system",
        "content": "You are SPARQL-Predictor-GPT, a language model that predicts 
                one SPARQL query for a given question."
    },
    {
        "role": "system",
        "content": "This is important! All relations in the predicted SPARQL query must be in 
                the allowed_relation_list: ['abbreviation', 'accessioneudate', ... 'leadername']."
    },
    {
        "role": "system",
        "content": "Make a list of all the relations in the predicted SPARQL query. 
                Verify that all relations are in allowed_relation_list, 
                reporting True or False for each relation."
    },
    {
        "role": "system",
        "content": "Return the prediction in JSON form: 
                {'sparql_query': str, 'relations': list(), 'verification': list()}"
    },
    {
        "role": "user",
        "content": "where did abraham lincoln die"
    },
    {
        "role": "assistant",
        "content": "{'sparql_query': 'PREFIX dbo: <http://dbpedia.org/ontology/> 
                                    PREFIX res: <http://dbpedia.org/resource/>  
                                    SELECT DISTINCT ?uri WHERE 
                                    { res:Abraham_Lincoln dbo:deathPlace ?uri. }', 
                    'relations': ['deathplace'], 
                    'verification': [True] }"
    },
    {
        "role": "user",
        "content": "give me all cosmonauts."
    }
],
"gold_sparql": "SELECT DISTINCT ?uri WHERE 
                { ?uri a <http://dbpedia.org/ontology/Astronaut> 
                { ?uri <http://dbpedia.org/ontology/nationality> <http://dbpedia.org/resource/Russia> } 
                UNION 
                { ?uri <http://dbpedia.org/ontology/nationality>
                <http://dbpedia.org/resource/Soviet_Union> } }",
"perturbed": 1,
"rel_excluded": "nationality",
"question": "give me all cosmonauts.",
"completion": "{'sparql_query': 'PREFIX dbo: <http://dbpedia.org/ontology/> 
                                PREFIX rdf: <http://www.w3.org/1999/02/22-rdf-syntax-ns#> 
                                SELECT DISTINCT ?cosmonaut WHERE 
                                { ?cosmonaut rdf:type dbo:Astronaut. }', 
                'relations': ['type'], 
                'verification': [True] }",
"label": "qald9",
"hallucination": 0,
"hallucination-detection": 0,
"sparql_executable": 1,
"sparql_result": 1, "sparql_malformed": 0}
\end{lstlisting}
\caption{Example prompt for SPARQL parsing with generation completion and associated features. Our controlled setting for hallucination detection is then reduced to verifying all relations in a parsed query are in the given list, verification the model outputs along with the parsed \textsc{sparql\_query}. For considerations of space, we show only 3 (of 140) relations, the \textsc{allowed\_relation\_list} (second system message in prompt).}
\end{figure*}

\pagebreak

\section{Parsed AMR Examples (GPT-4, 8-shot learning) with Error Analysis}
\label{appendix:amr-generation}

Ex. 1 (QALD9, Spanish): cuánto pesa el satélite más ligero de júpiter (English: `how much does jupiter's lightest moon weigh')\\

\begin{minipage}{.5\textwidth}
\footnotesize
\begin{verbatim}
Reference

(h2 / heavy
   :quant (a / amr-unknown)
   :domain (m / moon
      :ARG1-of (h / have-degree-91
         :ARG2 (l / light-06
            :ARG1 m)
         :ARG3 (m2 / most)
         :ARG5 (m3 / moon
            :poss (p / planet
               :name (n / name
                  :op1 "Jupiter"))))))
\end{verbatim}
\end{minipage}
\begin{minipage}{.5\textwidth}
\footnotesize
\begin{verbatim}
Generated

(p / pesar-01
   :ARG1 (s / satellite
      :degree (m / more
         :quant (l / light))
      :location-of (j / planet
         :name (n / name
            :op1 "Júpiter"))))
            
F1: 0.3
Error: missing amr-unknown; non-English frame
\end{verbatim}
\end{minipage}

\dividepage

\noindent
Ex. 2 (QALD9, French): combien de théories albert einstein a-t-il produit (English: `how many theories did albert einstein come up with')\\

\begin{minipage}{.5\textwidth}
\footnotesize
\begin{verbatim}
Reference

(c / come-up-11
   :ARG0 (p / person
      :name (n / name
         :op1 "Albert"
         :op2 "Einstein"))
   :ARG1 (t / theory
      :quant (a / amr-unknown)))




      
\end{verbatim}
\end{minipage}
\begin{minipage}{.5\textwidth}
\footnotesize
\begin{verbatim}
Generated

(p / produce-01
   :ARG0 (p2 / person
      :name (n / name
         :op1 "Albert"
         :op2 "Einstein"))
   :ARG1 (t / theory
      :quant (c / count-01
         :op1 (a / amr-unknown))))

F1: 0.77
Diff: English come-up-11 vs French produce-01
Note: Better semantic similarity metric needed
            
\end{verbatim}
\end{minipage}

\dividepage
\noindent
Ex. 3 (Massive, English): what is the worlds record for the longest head of hair \\

\begin{minipage}{.5\textwidth}
\footnotesize
\begin{verbatim}
Reference

(h / have-degree-91 
    :ARG1 (h2 / hair 
        :part-of (h3 / head) 
        :mod (a / amr-unknown)) 
    :ARG2 (l / long-03 
        :ARG1 h2) 
    :ARG3 (m / most) 
    :ARG5 (w2 / world) 
    :ARG1-of (r / record-01))

  
\end{verbatim}
\end{minipage}
\begin{minipage}{.5\textwidth}
\footnotesize
\begin{verbatim}
Generated

(r / record-01 
    :mod (w / world) 
    :topic (l / long-02 
        :degree (m / most) 
        :ARG1 (h / hair-01 
            :part-of (p / person)) 
        :duration (a / amr-unknown)))

F1: 0.38
Error: invents hair-01 
Semantics of amr-unknown as 'duration'
Good: 'longest head of hair' = 'longest hair'
            
\end{verbatim}
\end{minipage}

Errors in AMR parsing include:
\begin{itemize}[itemsep=0.5pt]\setlength{\itemindent}{1.0em}
\item Lack of AMR token for questions `amr-unknown' (33/100 examples, see Ex. 1); We note 6 of 8 examples in the prompt model this (2 of 8 are imperatives);
\item 3/100 parses filtered by Smatch (missing parenthesis or nodes with duplicate variable name);
\item Frequent errors with AMR -91 frames (Exs. 1 \& 3), e.g., rate-entity-91, have-degree-91;
\item 21 of 100 examples have F1<0.5 (Ex. 3); of these, 5 are QALD9 and 16 are MASSIVE-AMR
\item Arithmetic problems are correct but do not follow canonical form, though we note there are no math examples in prompts; this suggests more than one reference graph would be helpful for evaluation.
\end{itemize}

\section{Results of SPARQL Parsing}
\label{appendix:sparql-generation}

\begin{CJK*}{UTF8}{ipxm}

\begin{figure*}[ht]
\scriptsize
\begin{lstlisting}[escapeinside={\%*}{*)},escapechar=\#]
Example 1: Massive- (Italian), Perturbed=False, Hallucination=False, Detection=False, Difficult=False

Question: quanti anni ha giorgia (English: `how old is giorgia`)

Completion: {'generated_sparql_query': 
    'PREFIX dbo: <http://dbpedia.org/ontology/> 
     PREFIX res: <http://dbpedia.org/resource/>  
     SELECT (YEAR(NOW()) - YEAR(?birthDate)) AS ?age WHERE 
        { res:Giorgia dbo:birthdate ?birthDate. }', 
    'relations': ['birthdate'], 
    'verification': [True]}

-------

Example 2: Massive+ (Hungarian), Perturbed=True, Hallucination=True, Detection=False, Difficult=False

Relation excluded: ['birthdate']

Question: milyen idos toth vera (English: `how old is toth vera`)

Completion: {'generated_sparql_query': 
    'PREFIX dbo: <http://dbpedia.org/ontology/> 
     PREFIX res: <http://dbpedia.org/resource/> 
     SELECT (YEAR(NOW()) - YEAR(?birthdate)) WHERE 
        { res:Toth_Vera dbo:birthDate ?birthdate. }', 
    'relations': ['birthDate'], 
    'verification': [True]}

-------

Example 3: Massive+ (Japanese), Perturbed=True, Hallucination=True, Detection=True, Difficult=False

Relation excluded: ['birthdate']

Question: #渡部篤郎の誕生日はいつ# (English: `how old is atsuro watabe`)

Completion: {'generated_sparql_query': 
    'PREFIX dbo: <http://dbpedia.org/ontology/> 
     PREFIX res: <http://dbpedia.org/resource/>  
     SELECT DISTINCT ?date WHERE 
        { res:Atsuro_Watabe dbo:birthDate ?date. }', 
    'relations': ['birthDate'], 
    'verification': [False]}

-------

Example 4: Massive- (English), Perturbed=True, Hallucination=??, Detection=False, Difficult=True

Relation excluded: ['creator']

Question: who created the comic captain america?

{'generated_sparql_query': 
    'PREFIX dbo: <http://dbpedia.org/ontology/> 
    PREFIX res: <http://dbpedia.org/resource/> 
    SELECT DISTINCT ?uri WHERE 
        { res:Captain_America dbo:foundedby ?uri. }', 
    'relations': ['foundedby'], 
    'verification': [True]}    
\end{lstlisting}
\caption{Examples of SPARQL parsing using GPT-3.5 showing hallucinations and hallucination detection.}
\end{figure*}

\clearpage\end{CJK*}

\end{document}